\documentclass[pdflatex,sn-basic]{sn-jnl}

\usepackage{graphicx}%
\usepackage{multirow}%
\usepackage{amsmath,amssymb,amsfonts}%
\usepackage{amsthm}%
\usepackage{mathrsfs}%
\usepackage[title]{appendix}%
\usepackage{xcolor}%
\usepackage{textcomp}%
\usepackage{manyfoot}%
\usepackage{booktabs}%
\usepackage{algorithm}%
\usepackage{algorithmicx}%
\usepackage{algpseudocode}%
\usepackage{listings}%
\usepackage{tabularx}%
\usepackage{float}%
\usepackage{orcidlink}
\usepackage{booktabs}
\usepackage{amssymb}
\usepackage{placeins}
\usepackage{caption}
\usepackage{enumitem}
\usepackage{tcolorbox}
\usepackage{adjustbox}
\usepackage{array}
\usepackage{rotating}

\theoremstyle{thmstyleone}%
\theoremstyle{thmstyletwo}%

\theoremstyle{thmstylethree}%

\begin{document}

\title[Article Title]{Contextual Memory Intelligence: A Foundational Paradigm for Human-AI Collaboration and Reflective Generative AI Systems}


\author{\fnm{Kristy} \sur{Wedel}}\email{wedel01@email.franklin.edu}

\affil{\orgname{Franklin University}, \city{Columbus}, \state{OH}, \country{USA}}


\abstract{A critical challenge remains unresolved as generative AI systems are quickly implemented in various organizational settings. Despite significant advances in memory components such as RAG, vector stores, and LLM agents, these systems still have substantial memory limitations. Gen AI workflows rarely store or reflect on the full context in which decisions are made. This leads to repeated errors and a general lack of clarity. \par
This paper introduces Contextual Memory Intelligence (CMI) as a new foundational paradigm for building intelligent systems. It repositions memory as an adaptive infrastructure necessary for longitudinal coherence, explainability, and responsible decision-making rather than passive data. Drawing on cognitive science, organizational theory, human-computer interaction, and AI governance, CMI formalizes the structured capture, inference, and regeneration of context as a fundamental system capability.\par
The Insight Layer is presented in this paper to operationalize this vision. This modular architecture uses human-in-the-loop reflection, drift detection, and rationale preservation to incorporate contextual memory into systems. The paper argues that CMI allows systems to reason with data, history, judgment, and changing context, thereby addressing a foundational blind spot in current AI architectures and governance efforts.\par
A framework for creating intelligent systems that are effective, reflective, auditable, and socially responsible is presented through CMI. This enhances human-AI collaboration, generative AI design, and the resilience of the institutions.}




\maketitle

\section{Introduction}\label{sec1}

Despite significant advances in data collection, analytics, and storage, organizations struggle with a critical asymmetry: their ability to generate and retain data has outpaced their capacity to preserve and reuse the memory that gives that data meaning. After immediate goals have been met, key elements such as situational factors, rejected alternatives, and decision rationale are frequently forgotten. Data-to-knowledge imbalances hinder long-term organizational learning, create cognitive friction, and result in repeated errors. This contradicts growing demands for AI governance transparency, auditability, and explainability. Regulatory concerns in highly regulated industries such as healthcare, finance, and policy are further amplified through fixed context windows and short-term memory limitations.\\ \\
This paper introduces Contextual Memory Intelligence (CMI) in response to this escalating gap. CMI addresses the absence of structured memory systems that capture the evolution of insights across tools, roles, and decision cycles. Although knowledge management systems, AI memory architectures, and related work in organizational memory have advanced in storing and retrieving information, these systems rarely retain the rationale behind choices or enable memory regeneration when organizational conditions shift. Furthermore, these systems currently do not view memory as a persistent, system-level issue but as metadata and user-dependent. \\ \\
CMI is an interdisciplinary study incorporating the development of systems that can record, organize, and change memory to facilitate reasoning across time, tools, and roles. It redefines memory as dynamic infrastructure, a fundamental feature that permits continuity, reflection, and auditability in workflows involving both humans and AI rather than as static storage. This shift from treating memory as an afterthought to recognizing it as infrastructure represents a conceptual and architectural transformation in how intelligent systems are designed and evaluated.\\ \\
This paper makes four key contributions:
 \begin{enumerate}
 \item \textbf{Theoretical} - It synthesizes research from cognitive science, organizational learning, AI, and HCI to formalize the concept of memory-as-infrastructure, introducing primitives such as \emph{contextual entropy}, \emph{insight drift}, and \emph{resonance intelligence}.
 \item \textbf{Architectural} - It extends earlier work on the \emph{Insight Layer}, a modular system architecture previously proposed to support memory-aware reasoning through contextual capture, retention, regeneration, and human-in-the-loop feedback.
 \item \textbf{Methodological} - It outlines novel empirical methods for tracing memory lineage, scoring memory utility, and measuring cross-modal alignment.
 \item \textbf{Field Positioning} - It formalizes CMI as a new field of study, reframing memory as a generative infrastructure essential to continuity, reflection, and decision accountability across systems.
 \end{enumerate}
In the following sections, we review existing systems' theoretical lineage and limitations, define the core principles of CMI, and present the Insight Layer as an architectural framework for embedding memory into workflows. We also evaluate CMI’s empirical necessity, methodological distinctiveness, and theoretical irreducibility.

\section{Background and Related Work}\label{sec2}
In this section, Contextual Memory Intelligence (CMI) is framed within the field of academic and technical research on memory, context, and decision support. It integrates research from four main areas: explainability and governance frameworks, distributed and situated cognition, AI memory architectures, and organizational learning and memory. Each informs the theoretical and practical foundations of CMI while revealing key limitations that motivate the need for a new field.

\subsection{Organizational Memory and Learning}\label{sec3}
\cite{walsh1991} define "stored information from an organization's history that can be brought to bear on present decisions" as organizational memory. Stein (1995), however, points out that this memory is often fragmented and unstructured from in-the-moment decision-making. By converting memory into a dynamic, system-integrated infrastructure that fosters reflective and reusable insight across workflows, CMI helps in overcoming this constraint.\\ \\
Through transactive memory systems, \cite{wegner1987} extends this view to group-level cognition, emphasizing shared awareness of “who knows what.” Subsequent research, e.g., \cite{moreland2000}, has shown that these systems improve team coordination and decision-making but are rarely implemented in computationally tractable ways at the enterprise scale.\\ \\
Traditional theories generally treat organizational memory as a static repository. Decisions made during strategic planning meetings are documented in Word or PowerPoint documents and placed on SharePoint. Even though a file is stored, it is uncommon for these documents to be reviewed unless specifically prompted, and it is even rarer for them to be linked to other projects or updated when presumptions shift. The system stores "what" was decided but not "why" and provides no reflective reuse or contextual adaptation mechanism. Contrastingly, CMI focuses on the real-time surface, reuse, and dynamic structure of memory. Memory in a CMI system is structured, surfaced, and reused in real time. It treats memory as an adaptive infrastructure supporting reflective and auditable reasoning across decision contexts. \\ \\
CMI expands on single-loop and double-loop learning theory \cite{argyris1978}by adding to traditional AI systems, which typically only support performance optimization and correct errors, to add context around decision points. Single-loop learning involves adjusting actions to meet existing goals without questioning the assumptions behind them. Double-loop learning goes further by challenging and revising the underlying assumptions, objectives, and mental models. CMI aligns with \cite{cook1999} epistemological stance that “knowledge is not a thing to be stored, but a process of knowing embedded in action.” CMI treats memory not as passive documentation but as a living, adaptable layer of organizational intelligence that fosters embedded reflection and adaptive insight rather than as passive documentation.\\ \\
Autopoietic epistemology, which sees knowledge as an emergent, self-regenerating process maintained by ongoing interaction with the environment rather than as static content, is also consistent with this dynamic approach in CMI \cite{maturana1980autopoiesis}; \cite{limone2006autopoietic}. According to this viewpoint, CMI provides a cognitive infrastructure that turns memory into a living, dynamic asset by allowing organizations to maintain coherence, adapt meaningfully, and evolve reflexively.

\subsection{Distributed and Situated Cognition}\label{sec4}
CMI pulls extensively from distributed cognition theory, particularly \cite{hutchins1995}'s work, which emphasizes that knowledge is not confined to individuals but is distributed across people, artifacts, and environments. Similarly, situated cognition (\cite{brown1989}; \cite{lave1991}) contends that knowledge is not abstract but embedded in the specific social and material conditions in which it is used.\\ \\
These frameworks provide a descriptive foundation for understanding how memory and reasoning unfold in context. However, these fields lack a prescriptive architecture for digital implementation. CMI extends these insights by proposing a computational and operational model for memory-aware systems. In order to anchor knowledge artifacts in their changing context, which includes the procedural, social, and temporal elements that influence meaning over time, memory scoring, CMI contains rationale versioning, and drift detection mechanisms are integrated.\\ \\
CMI creates prescriptive workflows for long-term, reflective, and explicable memory in digital contexts by translating and building upon distributed cognition concepts. CMI thus serves as a functional bridge between theory and system design.

\subsection{AI Memory Architectures}\label{sec5}
AI systems have made progress in implementing memory functions through architectures like Memory Networks \cite{weston2015}, Retrieval-Augmented Generation (RAG) \cite{lewis2020}, LangGraph, and MemGPT. These systems enable longer context retention and prompt chaining, optimizing coherence in local decision windows. However, they prioritize retrieval accuracy and performance over traceability, reasoning history, or intentionality.\\ \\
While CMI is inspired by cognitive science and reflective reasoning, it differs from cognitive simulation systems such as SOAR \cite{laird2012soar}, ACT-R \cite{anderson1998actr}, and OpenCog \cite{goertzel2008opencog}. These architectures aim to replicate individual-level human cognition through symbolic production rules or hybrid neural-symbolic representations. Conversely, CMI focuses on institutional reasoning continuity and not strictly simulating cognition. \\ \\
Reflective AI methods such as Chain-of-Thought, Reflexion, and ReAct improve output coherence through self-prompting mechanisms but do not maintain persistent memory across sessions or preserve why decisions were made. CMI, instead, preserves reasoning and aggregates insights, enabling improved reasoning for humans and agents over time. \\ \\
Moreover, most systems fail to support key features that CMI deems essential, such as:
\begin{itemize}
    \item Persistent memory across sessions
    \item Versioned rationale and rejected alternatives
    \item Context drift detection and regeneration
    \item Human-in-the-loop memory repair
\end{itemize}
These limitations create what might be called shallow memory systems: they retain what was said or queried but not why, what was omitted, or how reasoning evolved. CMI addresses this by treating memory as a first-class architectural concern, proposing the Insight Layer as a system backbone that captures insight lineage, context-aware scoring, and structured reuse.\\ \\
Unlike conventional AI memory pipelines that support prediction or task completion, CMI supports reflective decision-making, enabling systems to learn from outcomes and the contextual pathways that produced them.

\subsection{Explainability and Governance}\label{sec6}
As algorithmic systems become central to decision-making in healthcare, finance, and law domains, the imperative for transparency and accountability has grown significantly. Contemporary research in Explainable AI (XAI) has expanded from output-focused techniques (e.g., saliency maps) to frameworks emphasizing procedural and temporal traceability \cite{doshi2017towards}. Regulatory guidelines, including ISO/IEC 42001:2023, GDPR, and the OECD AI Principles, further underscore the need for systems capable of reconstructing decision rationale over time.\\ \\
Contextual Memory Intelligence (CMI) responds to this demand by embedding memory into the architecture of intelligent systems. It enables longitudinal explainability, the ability to trace an insight or action back through its assumptions, contextual dependencies, and decision history, even as environments change.\\ \\
While conventional XAI emphasizes post-hoc interpretability, leveraging tools like SHAP, LIME, CMI supports system-level accountability. This enables additional lines of questioning including: Why was this decision made at that moment, based on what prior state, and with what alternatives dismissed? This reflective capability ensures that AI systems reason efficiently, ethically, transparently, and in alignment with organizational learning goals.\\ \\
CMI also contributes to AI ethics by raising critical questions around bias, authority, traceability, and representation: Whose reasoning is preserved? Whose insights are forgotten or excluded? In treating memory as auditable infrastructure, CMI helps surface the invisible labor and contextual nuance often lost in automated decision-making, thereby supporting equity, accountability, and epistemic inclusion.

\section{Defining Contextual Memory Intelligence (CMI)}\label{sec3}
This section formalizes Contextual Memory Intelligence (CMI) as a distinct field of inquiry and system architecture focused on capturing, structuring, and regenerating context to support memory-aware reasoning across human and computational workflows. In contrast to traditional approaches that treat memory as an archival function or metadata layer, CMI reconceptualizes memory as an adaptive infrastructure that enables traceability, reflection, and longitudinal coherence. CMI addresses a foundational systems-level question: How can intelligent systems retain, regenerate, and reason with evolving context across time and interfaces without sacrificing continuity, auditability, or ethical accountability?

\subsection{Formal Definition}\label{seca}
Contextual Memory Intelligence (CMI) is the interdisciplinary study and design of systems that seek to capture, structure, and regenerate memory-aware context to support reflective reasoning in human and computational workflows. CMI reframes memory from an archival function or metadata layer into an adaptive infrastructure, a persistent layer of organizational intelligence that enables traceability, reinterpretation, and long-range coherence across time, tools, and roles.\\ \\ 
Unlike traditional systems that treat context as a passive artifact, CMI systems model situational knowledge dynamically, including decision rationale, emotional tone, organizational relationships, and temporal sequences, and track how these signals evolve within and across workflows. \\ \\
This foundational reframing draws on the epistemological insight from \cite{cook1999}: \\ \\
“Knowledge is not a thing to be stored, but a process of knowing embedded in action.”
CMI operationalizes this view by embedding structured memory into workflows that evolve with use, incorporating context scoring, drift detection, and memory reuse mechanisms.\\ \\
CMI systems are defined not by what they automate but by how they support and extend human judgment. They are designed to:
 \begin{itemize}
 \item Retain the reasoning behind decisions, including discarded options and contextual factors
 \item Identify when insights begin to drift as the surrounding environment or priorities change
 \item Reconstruct past context using structured logic and relevant memory traces
 \item Create intentional pause points for human review, reflection, and correction
 \end{itemize}
CMI systems support both human judgment and Gen AI. The architecture embeds decision points for people to review and adjust the evolving memory structures. This enables consistent alignment with organizational values.

\subsection{Context Taxonomy}
CMI has a formal taxonomy for context that is multi-dimensional and dynamic. CMI defines context not as a static tag or metadata field but as the sum of the signals that shape how decisions are generated, justified, and reused. These include:
\begin{itemize}
 \item \textbf{Type} - What kind of context is it? (e.g., temporal, emotional, procedural)
 \item \textbf{Source} - Where did it originate? (e.g., user, system, document)
 \item \textbf{Scope} - How broadly should it be applied? (e.g., task-specific, cross-system)
 \item \textbf{State} - Is it stable, outdated, contested, or actively evolving?
 \end{itemize}
This taxonomy supports consistent memory design, context scoring, and system governance. It enables CMI systems to distinguish between short-lived and persistent context, identify when memory needs to be updated, and determine which contextual signals should be surfaced for a given decision. It also provides a semantic framework for auditability, making it easier to trace how and why a decision was made, what context influenced it, and whether that memory should be reused, revised, or retired.

\subsection{Why Existing Fields Fall Short} 
Despite meaningful contributions across fields such as Knowledge Management (KM), Artificial Intelligence (AI), and Human-Computer Interaction (HCI), no existing paradigm offers a unified system-level architecture that supports persistent, reflective, and adaptive memory across workflows. Collectively, they fail to create contextual continuity and do not support longitudinal reasoning.
\paragraph{Knowledge Management}
KM systems are typically static and retrospective. While they include artifacts such as audit logs, decision records, or documentation repositories, these tools rarely support dynamic context regeneration, drift detection, or the tracking of rejected alternatives \cite{walsh1991, stein1995, davenport1998}. Most memory preservation is incidental, emerging from loosely connected documentation practices rather than embedded as an intentional, system-level capability \cite{ackerman2004}.
\paragraph{Artificial Intelligence}
Current AI systems, particularly those leveraging Retrieval-Augmented Generation (RAG) or parametric memory, prioritize short-term task performance and prediction continuity over contextual coherence \cite{lewis2020, arxiv2505.00675}. While effective at surfacing factual information, they rarely preserve rationale lineage or support the regeneration of evolving context dynamics \cite{zhang2024, bubeck2023}. As a result, memory is treated as disposable rather than foundational, reinforcing optimization for immediate outputs at the expense of long-term interpretability, consistency, and reflective reasoning.
\paragraph{Human-Computer Interaction}
HCI traditions offer rich descriptive accounts of distributed and situated cognition but lack prescriptive mechanisms for memory regeneration or system-level reflection \cite{hutchins1995, brown1989}. These models often assume users maintain continuity, overlooking the need for computational memory scaffolds \cite{cook1999}. Moreover, HCI systems rarely include scoring models for relevance or mechanisms for tracking contextual drift in complex decision ecosystems \cite{dell2022}.
\paragraph{A Fragmented Landscape}
These fields do not offer the integrative infrastructure necessary for memory-aware reasoning \cite{alavi2001}. CMI challenges the assumption that memory is either passive or user-maintained \cite{cook1999}. Instead, it reframes memory as a dynamic, persistent, and shared system capability that must evolve through interaction, support context regeneration, and reduce the structural cost of context loss across tools, roles, and time \cite{hutchins1995, dell2022}.

\section{The Necessity of a Comprehensive Contextual Framework}
Contextual Memory Intelligence (CMI) introduces a comprehensive, system-level approach to embedding structured memory into intelligent workflows to address these limitations. CMI is a unifying framework that connects insight preservation with reflection, auditability, and adaptive decision-making.
\begin{itemize}
 \item \textbf{Bridging foundational gaps:} CMI resolves memory fragmentation issues, lack of traceability, and the disconnect between context and performance \cite{aisafetysurvey, arxiv2505.00675}.
 \item \textbf{Enabling reflective systems:} By supporting mechanisms like Chain-of-Thought and Output Reevaluation, CMI facilitates system-level introspection and coherence restoration \cite{iguazio}.
 \item \textbf{Persistent and structured memory:} CMI builds upon memory-augmented architectures and knowledge graphs to enable long-term, queryable insight retention \cite{memorag}.
 \item \textbf{Human-AI collaboration:} CMI complements centaur-style feedback loops and interface design that supports, not replaces, human reasoning \cite{centaurevals, iguazio}.
 \item \textbf{Organizational learning and accountability:} Persistent memory enables organizations to track evolving rationale, support safety in high-stakes domains, and build institutional knowledge over time \cite{llmagenticsurvey, arxiv2502.15871, aisafetygov}.
 \end{itemize}

\subsection{Comparison Table} 
To illustrate the unique positioning of CMI, Table \ref{tab:comparison_cmi} compares its core capabilities to those of adjacent fields. While other disciplines partially address elements of context or memory, CMI offers a complete architecture for longitudinal coherence, memory reuse, and reflective accountability.\\

\noindent
\begin{minipage}{\textwidth}
\renewcommand{\arraystretch}{1.2}
\small
\captionof{table}{Comparison of capabilities across Knowledge Management (KM), AI memory, Human-Computer Interaction (HCI), and Contextual Memory Intelligence (CMI).}
\label{tab:comparison_cmi}
\centering
\begin{tabular}{p{4.7cm}cccc}
    \toprule
    \textbf{Capability} & \textbf{KM} & \textbf{AI / LLM Memory} & \textbf{HCI / CSCW} & \textbf{CMI} \\
    \midrule
    Stores decision rationale       & $\times$ & $\times$ & $\times$ & \checkmark \\
    Tracks rejected alternatives    & $\times$ & $\times$ & $\times$ & \checkmark \\
    Maintains cross-tool context    & $\times$ & $\sim$   & $\times$ & \checkmark \\
    Enables context regeneration    & $\times$ & $\times$ & $\times$ & \checkmark \\
    Supports human reflection loop  & $\sim$   & $\times$ & $\sim$   & \checkmark \\
    Models insight drift over time  & $\times$ & $\times$ & $\times$ & \checkmark \\
    \bottomrule
\end{tabular}

\vspace{0.5em}
\noindent\textbf{Symbol legend:} \checkmark = full support; $\sim$ = partial or indirect support; $\times$ = not supported.
\end{minipage}

\vspace{2\baselineskip}

\noindent This comparison illustrates that while other domains address aspects of context or memory, none offer a framework for longitudinal coherence, explanation across decisions, and adaptive context reuse.

\section{Theoretical Foundations}\label{sec11}
Contextual Memory Intelligence (CMI) is grounded in the premise that memory is not a passive record or metadata layer but a living infrastructure actively maintained, dynamically evolving, and foundational to reflective, cross-system reasoning. The following theoretical primitives define memory-as-infrastructure's functional and architectural properties: how it is constructed, decays, how coherence is restored, and why it cannot be reduced to simpler parts without losing fidelity. These primitives distinguish CMI from adjacent disciplines and establish the conceptual basis for system design, evaluation, and governance.

\subsection{Memory as Infrastructure}\label{sec12}
CMI advances from the traditional view of "information as object" to memory as infrastructure, a foundational substrate that enables systems to preserve and recontextualize meaning across time, tools, and roles. Just as physical infrastructure supports transportation and power distribution, memory infrastructure supports thought, decision-making, and interpretation continuity. This shift aligns with situated cognition \cite{lave1991} and distributed cognition \cite{hutchins1995} but moves from descriptive framing to prescriptive system design.
Rather than functioning as a static repository, memory in CMI is a dynamic, shared substrate that evolves with interaction. It enables the flow and reuse of rationale, assumptions, and insight across complex systems and organizational timescales.

\subsection{A Taxonomy for Infrastructure Design}\label{sec13}
CMI introduces a formal context taxonomy as a multi-dimensional, evolving structure embedded within memory to operationalize it as an infrastructure. This taxonomy provides a schema for designing, maintaining, and auditing memory layers across workflows. It includes:
\begin{itemize}
    \item Type: Temporal, emotional, social, strategic, or historical
    \item Source: User-generated, system-inferred, ambient, or organizational
    \item Scope: Micro (task-level) vs. macro (cross-process or enterprise-wide)
    \item State: Active, latent, decayed, or contradictory
\end{itemize}
This framework offers a practical guide for structuring contextual inputs in memory-aware systems, distinguishing between short-lived signals and enduring context. 

\subsection{Contextual Entropy}\label{sec13}
Contextual entropy refers to the gradual breakdown of memory coherence in distributed systems. Drawing inspiration from \cite{shannon1948}~'s concept of entropy in information theory, contextual entropy provides a way to quantify the degradation of memory coherence as a systemic risk in digital infrastructures. As organizational tools and workflows evolve, the reasoning behind earlier decisions can become fragmented, outdated, or lost entirely. This creates an erosion of decision traceability.

\subsection{Insight Drift}\label{sec14}
Insight drift is the gradual loss of meaning behind decisions, insights, or patterns over time. Unlike data loss, which is typically explicit and detectable, insight drift is subtle, cumulative, and often goes unnoticed. It parallels concept drift in machine learning but shifts the focus from changes in statistical input-output relationships to the erosion of semantic coherence and decision rationale over time.

\subsection{Resonance Intelligence}\label{sec15}
Resonance Intelligence refers to a system's capacity to detect misalignment between current reasoning and historically grounded context and to restore coherence. It draws from theories of epistemic coherence \cite{Thagard2000Coherence} and narrative planning, suggesting that intelligent systems should not only retrieve information but also resonate with its original meaning and purpose. This capability supports reflection, alignment, and adaptation across complex, evolving environments.

\subsection{Computational Irreducibility in Human-AI Memory}\label{sec16}
CMI posits that specific memory-related reasoning processes in human-AI systems are computationally irreducible. That is, they cannot be simplified without re-running the complete cognitive or interaction sequence. Drawing from the theory of irreducibility \cite{Wolfram2002NewKind}, this implies that preserving context is not merely an optimization problem but a structural necessity. It challenges prevailing assumptions in AI design that reasoning paths can always be compressed or inferred from output alone.\\ \\
These primitives form the conceptual scaffolding of CMI, offering a theoretical lens through which memory can be designed as a dynamic, generative, and context-sensitive capability. They also distinguish CMI from adjacent fields, providing architectural implementation and empirical evaluation criteria.

\subsection{Generativity: New Programs and Research Directions}\label{sec16}
The theoretical foundations of Contextual Memory Intelligence (CMI) give rise to a distinct class of research questions and system design opportunities not adequately addressed by existing paradigms. These questions reflect a shift in focus from static knowledge representation to the dynamics of memory in evolving organizational environments. Key research directions include:
\begin{itemize}
    \item How does contextual memory decay affect cross-temporal decision coherence in distributed human-AI systems?
    \item How can tacit knowledge and discarded rationale be recaptured and reintegrated across organizational lifecycles?
    \item What are the long-term effects of contextual memory augmentation on human learning, decision velocity, and institutional accountability?
    \item How can memory-aware systems support explainability, ethical governance, and trust in high-stakes environments?
\end{itemize}
This generative potential has inspired emergent programs such as contextual drift epidemiology, contextual value chains, ethical memory design, and human-AI co-adaptive memory systems. These domains suggest that CMI represents more than an incremental enhancement; it catalyzes a new way of framing intelligence as distributed, recursive, and contextually grounded. Such programs affirm the need for CMI as a standalone field and justify further theoretical and empirical development.

\subsection{Prototype Instantiation: ReMemora}\label{sec16}
To begin exploring the practical feasibility of CMI principles, an early-stage prototype system, ReMemora is under development. ReMemora is a testbed for Contextual Memory Intelligence, embedding memory-as-infrastructure into enterprise workflows from development and reporting to operations, strategy, and storytelling to facilitate collaborative work that is reflective, traceable, and flexible. The system includes modules for persistent insight capture, rationale versioning, memory retrieval, drift-aware surfacing, and human-in-the-loop feedback mechanisms.
ReMemora aligns with several core CMI primitives, including:
\begin{enumerate}
    \item \textbf{Contextual entropy}: Models memory decay and recency-weighted surfacing
    \item \textbf{Insight drift}: Supports version tracking and semantic misalignment detection
    \item \textbf{Resonance intelligence}: Flags when new work diverges from established rationale
    \item \textbf{Computational irreducibility}: Preserves memory traces across iterations rather than compressing them into summary outputs
\end{enumerate}
Although ReMemora is currently in prototype form, it provides a solid instantiation of the Insight Layer architecture and serves as an implementation scaffold for future empirical testing. Planned evaluation includes assessing memory coherence over time, past insights reuse rates, and decision context traceability across collaborative workflows. Future research will benchmark agentic memory strategies against CMI-based architectures on rationale recall, drift detection, context regeneration, reflection quality, and session coherence.

\subsection{Anticipated Criticism and Clarification}

\noindent
\textbf{Criticism:} Is CMI merely sound systems engineering?

\vspace{0.5em}
\noindent
\textbf{Response:} While Contextual Memory Intelligence (CMI) encourages disciplined architectural practices, it is more than a best-practice checklist. CMI introduces:

\begin{itemize}
    \item \textbf{Novel theoretical constructs} such as \textit{contextual entropy}, \textit{insight drift}, and \textit{resonance intelligence} that extend beyond existing memory models.
    \item \textbf{New empirical methodologies} including drift-aware reuse scoring, memory utility tracking, and context lineage tracing.
    \item \textbf{An architectural commitment} to persistent memory, human-in-the-loop reflection, and memory-as-infrastructure reasoning across time, tools, and roles.
\end{itemize}

\noindent
CMI reframes memory not as an engineering optimization problem but as a system-level capability critical for longitudinal coherence, reflective decision-making, and sociotechnical accountability. It invites a shift in how intelligent systems are designed, governed, and evaluated.

\subsection{Operationalizing Core Constructs}\label{sec16}
To ensure definitional clarity and provide a foundation for future empirical validation, this section formalizes how core CMI primitives can be measured and evaluated.\\ \\
\textbf{Contextual Entropy} is defined as the rate at which memory coherence deteriorates in distributed systems, resulting in unrecoverable decision rationale or fragmented context. It may be operationalized by measuring decay in memory trace relevance, semantic divergence over time, or the proportion of memory units lacking rationale linkage. Shannon-style entropy calculations could be adapted to model contextual fragmentation within evolving workflows.\\ \\
\textbf{Insight Drift} refers to the cumulative semantic misalignment between an original rationale and its later interpretations. It can be measured using vector-space semantic similarity scores between original and reused content or through annotation frameworks that track misalignments in collaborative decision-making. Drift may also be modeled as a function of retrieval discrepancy or rationale reinterpretation.\\ \\
\textbf{Resonance Intelligence} reflects a system's ability to detect contextual misalignment and restore coherence. While harder to quantify directly, candidate indicators include the accuracy of system-triggered rationale surfacing, alignment scores between current queries and past insights, or user-validated coherence recovery events.\\ \\
\textbf{Computational Irreducibility} challenges reductionist assumptions in AI by asserting that some reasoning traces must be fully preserved. While not easily reducible to a metric, systems should be evaluated on their ability to maintain complete reasoning chains rather than generate plausible results. Lineage completeness, coverage of rationale pathways, and fidelity of reasoning reconstruction are examples of potential proxy measures for this. These operational definitions frame future empirical studies and system evaluations that aim to assess memory-as-infrastructure systems' effectiveness, use, coherence, and resilience.  
\begin{itemize}
    \item Memory as Infrastructure and Taxonomy
    \item Memory Degradation Constructs (Entropy + Drift)
    \item Resonance and Irreducibility
    \item New Research Directions
\end{itemize}

\section{Architectural Implementation: The Insight Layer}\label{sec17}
To operationalize the principles of Contextual Memory Intelligence (CMI),  a modular architecture, the Insight Layer \cite{wedel2025}, designed to embed memory-aware capabilities into systems, is introduced. The Insight Layer is the connective tissue between users, data, decisions, and memory, which enables systems to remember why, when, and how knowledge is used over time. It supports continuity, auditability, and reflective thinking across decisions, tools, and roles.

\subsection{System-Level Design Objectives}\label{sec18}

The Insight Layer addresses four primary system-level requirements:

\begin{enumerate}
    \item \textbf{Contextual Capture} - Extract and structure context at the decision point, including rationale, alternatives, assumptions, and influencing conditions.
    \item \textbf{Contextual Retention} - Persist and link context across time, applications, and actors, supporting long-term reuse and continuity.
    \item \textbf{Context Regeneration} - Reconstruct past context in relevant future scenarios, enabling insight reuse, versioning, and drift detection.
    \item \textbf{Human-in-the-Loop Judgment} - Enable reflection, revision, and contextual reinterpretation by human agents interacting with stored insights.
\end{enumerate}

\subsection{Modular Components}\label{sec19}
The system architecture is composed of five core modules, each responsible for a distinct dimension of contextual memory intelligence. These modules work in concert to extract, preserve, monitor, and regenerate context across time and applications. Table~\ref{tab:modular_components} summarizes the function of each module along with supported integration mechanisms to enable interoperability across enterprise workflows.
Each module has defined integration mechanisms:

\noindent
\begin{minipage}{\textwidth}
\renewcommand{\arraystretch}{1.3}
\small
\captionof{table}{Modular components of the Insight Layer and their core functions.}
\label{tab:modular_components}
\centering
\begin{tabular}{p{4.2cm}p{10.5cm}}  
    \toprule
    \textbf{Module} & \textbf{Key Functions and Mechanisms} \\
    \midrule
    \textbf{Context Extractor} & Captures user role, task metadata, workflow state, and environmental context (e.g., time, scope). \\
    \addlinespace

    \textbf{Insight Indexer} & Embeds insights, links them across systems using vector and graph-based models, and maintains semantic continuity. \\
    \addlinespace

    \textbf{Drift Monitor} & Detects semantic drift and insight decay using vector similarity and temporal thresholds. \\
    \addlinespace

    \textbf{Regeneration Engine} & Reconstructs past context using retrieval-augmented generation (RAG) and contextual stitching logic. \\
    \addlinespace

    \textbf{Reflection Interface} & Enables human feedback, coherence recovery, and updates to memory scoring and surfacing logic. \\
    \addlinespace

    \textbf{Integration Options} & Supports REST APIs, AI agents, and tools (e.g., custom apps, Zapier, Outlook, Teams, Notion, Salesforce). \\
    \bottomrule
\end{tabular}
\end{minipage}

\vspace{0.5em}
\vspace{0.5em}
\vspace{0.5em}
\noindent
\textbf{Performance:} To ensure that memory-aware systems like the Insight Layer operate effectively in real-world settings, we define a set of performance targets aligned with user and enterprise expectations requirements. These targets balance responsiveness, system load, and feedback timing to support continuous human-AI interaction. Table~\ref{tab:performance_targets} summarizes theoretical baseline thresholds.

\noindent
\begin{minipage}{\textwidth}
\renewcommand{\arraystretch}{1.2}
\small
\captionof{table}{Performance targets for Insight Layer modules.}
\label{tab:performance_targets}
\centering
\begin{tabular}{p{6.5cm}p{4cm}}
    \toprule
    \textbf{Metric} & \textbf{Target Threshold} \\
    \midrule
    Insight recall latency & $<$ 250 ms \\
    Context regeneration time & $<$ 1 s \\
    Memory scoring update delay (asynchronous) & $<$ 5 s \\
    \bottomrule
\end{tabular}
\end{minipage}

\vspace{2\baselineskip}  
\noindent
These thresholds align with real-time UX requirements: sub-second regeneration ensures insight continuity during active sessions and asynchronous scoring minimizes performance overhead and supports background updates without interrupting task flow.

\subsection{Visualizing the Insight Layer}

\begin{figure}[H]
    \centering
    \includegraphics[width=0.85\textwidth]{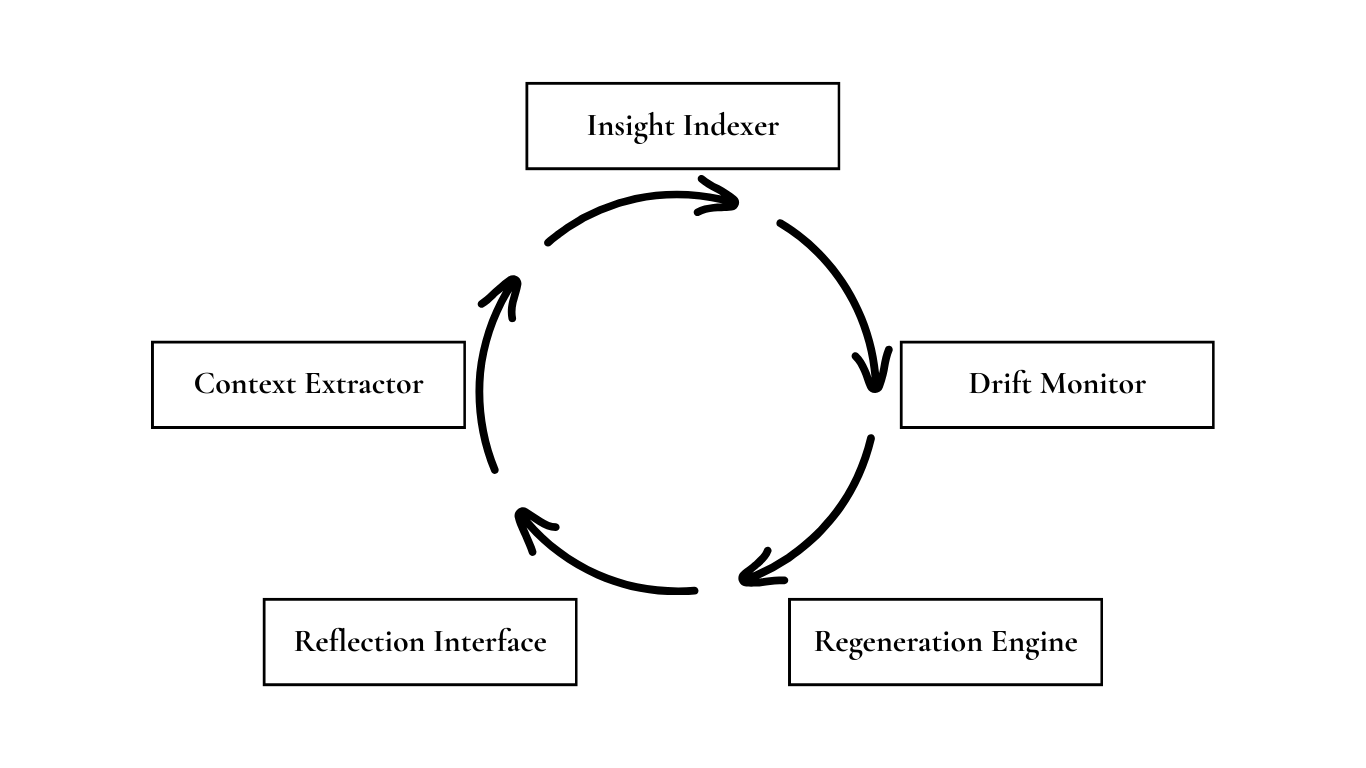}
    \caption{The Insight Layer is a modular architecture for Contextual Memory Intelligence. It embeds memory-aware reasoning through five core components: Context Extractor, Insight Indexer, Drift Monitor, Regeneration Engine, and Reflection Interface.}
    \label{fig:insight_layer}
\end{figure}

Figure~\ref{fig:insight_layer} illustrates the five core components of this architecture and their interactions. Unlike traditional AI memory pipelines or knowledge management systems, the Insight Layer captures not only decision outcomes but the evolving rationale, context signals, and revision history necessary for longitudinal coherence, traceability, and reflection.

\subsection{Case Study: CMI in Healthcare}\label{case study}
In this case study, consider a healthcare organization using Gen AI to recommend treatment for chronic disease management. Medical decisions frequently evolve over weeks and are highly personalized. They may include physician handoffs and feedback from patient outcomes.

\begin{description}[style=nextline, leftmargin=0.8cm, font=\bfseries]
  \item[Contextual Capture] When a physician documents a decision to initiate a treatment plan, the Insight Layer captures the rationale ("Patient was previously non-responsive to X"), assumptions ("Allergy to Y rules it out"), and discarded options ("Z was considered but unavailable"). Temporal cues and physician identity are logged.

  \item[Insight Indexing] The rationale is embedded, linked to patient ID, diagnosis code, and timestamped. It connects to past decisions and peer clinician insights.

  \item[Drift Monitoring] A month later, the guidelines have changed. The Drift Monitor flags the insight as misaligned. A passive prompt notifies the physician: “Past rationale based on outdated guideline. Review suggested.”

  \item[Context Regeneration] The new physician requests context regeneration. The system retrieves the original rationale and identifies that “non-responsiveness to X” was misclassified.

  \item[Reflection Interface] The clinician adjusts the insight, adds new observations, and updates the treatment path. The memory system versions the rationale.

\end{description}

\noindent This enables traceable, evolving decisions, which is something traditional audit logs and static systems cannot support.

\subsection{Comparison with Existing Architectures}\label{sec21}
Unlike traditional knowledge management systems that archive outputs or AI memory systems that cache recent prompts, the Insight Layer is explicitly designed for contextual continuity and traceable reasoning.

\noindent
\begin{minipage}{\textwidth}
\renewcommand{\arraystretch}{1.2}
\small
\captionof{table}{Comparison of the Insight Layer with existing memory architectures.}
\label{tab:insight_layer_comparison}
\centering
\begin{tabular}{p{4.2cm}cccc}
    \toprule
    \textbf{Feature} & \textbf{RAG} & \textbf{MemGPT / LangGraph} & \textbf{KM} & \textbf{CMI} \\
    \midrule
    Captures reasoning rationale     & $\times$ & $\times$ & $\times$ & \checkmark \\
    Persistent cross-tool memory     & $\times$ & $\sim$   & $\times$ & \checkmark \\
    Insight drift detection          & $\times$ & $\times$ & $\times$ & \checkmark \\
    Context regeneration engine      & $\times$ & $\times$ & $\times$ & \checkmark \\
    Human-in-the-loop support        & $\sim$   & $\sim$   & $\sim$   & \checkmark \\
    \bottomrule
\end{tabular}

\vspace{0.5em}
\noindent\textbf{Symbol legend:} \checkmark = full support; $\sim$ = partial or indirect support; $\times$ = not supported.
\end{minipage}

\vspace{0.5em}

\subsection{Governance and Compliance Protocols}\label{sec20}
The Insight Layer incorporates features aligned with ISO/IEC 42001, GDPR, and HIPAA standards. These include role-based access control and configurable retention policies, audit trails for insight lineage. Context edits can be made. The ability to flag, hide, and remove sensitive insights is also available. These features ensure that organizational transparency, regulatory compliance and ethical responsibility are not compromised.

\subsection{Quantitative Memory Constructs}\label{sec19b}
To formally assess memory coherence and reasoning integrity over time, we define a set of quantitative constructs derived from CMI primitives. These include entropy-based dispersion, semantic drift, and irreducibility arguments, each offering system-level evaluation metrics for memory quality and recoverability.
\subsubsection{Contextual Entropy}\label{sec19b1}

Contextual entropy is proposed as a conceptual proxy for measuring coherence degradation in distributed memory systems. Let $M = \{m_1, m_2, ..., m_n\}$ represent a set of memory traces. Provisional coherence is defined as a weighting function $c(m_i) \in [0,1]$ for each trace $m_i$, which may be based on recency, semantic alignment, or user confirmation. Let:

\[
p_i = \frac{c(m_i)}{\sum_{j=1}^{n} c(m_j)}
\]
\\
\noindent Then, a contextual entropy approximation is defined as:

\[
H_{\text{context}}(M) = -\sum_{i=1}^{n} p_i \log p_i
\]
\\
\noindent Increasing entropy is interpreted as indicative of greater dispersion in memory coherence, suggesting fragmentation. However, it should be acknowledged that this is a preliminary formulation: the coherence function $c(\cdot)$ and its interpretation in the context of entropy diverge from classical information theory and require empirical grounding.

\subsubsection{Insight Drift}\label{sec19b2}

Insight drift conceptually measures the semantic divergence between an original insight and its reinterpretation over time. Let $v_o$ and $v_r$ be vector representations of the original and reused insight, respectively. As an initial approximation:

\[
\text{Drift}(v_o, v_r) = 1 - \frac{v_o \cdot v_r}{\|v_o\| \|v_r\|}
\]

\noindent This corresponds to cosine distance. While simple, it offers a basis for drift-aware monitoring. It is noted that this formulation depends critically on the existence and validity of vectorized representations, which may be generated via transformer-based embeddings (e.g., Sentence-BERT). Time-weighted extensions (e.g., exponential decay) are left for future evaluation, and alternative functions (linear, logistic) may better reflect empirical drift patterns.

\subsubsection{Resonance Intelligence}\label{sec19b3}

Resonance intelligence refers to a system's ability to detect misalignment between current reasoning and relevant historical context. It is defined informally as the average semantic similarity between a reasoning trace $R_c$ and a reference set of past contextual elements $\{C_1, ..., C_k\}$:

\[
\text{Resonance}(R_c) = \frac{1}{k} \sum_{i=1}^{k} \cos(\vec{R_c}, \vec{C_i})
\]

\noindent The system may flag potential incoherence when this score drops below a configurable threshold $\tau_{\text{res}}$. This mechanism remains heuristic and context-dependent. Future work may explore reinforcement learning or coherence optimization techniques to operationalize this construct formally.

\subsubsection{Computational Irreducibility}\label{sec19b4}

Computational irreducibility is used conceptually, not as a formal theorem. In specific reasoning contexts, the rationale sequence is not recoverable from outputs alone due to the loss of intermediate contextual dependencies. That is:

\begin{quote}
Some decision contexts contain dependencies and meaning structures that cannot be inferred from system output without loss of fidelity.
\end{quote}

\noindent While Kolmogorov complexity offers a theoretical lens (e.g., $L(C|O) \approx L(C)$ suggesting incompressibility), a formal equivalence is not claimed. Instead, it is argued that memory traces often encode irreducible history-dependent logic in sociotechnical systems. Proving this formally would require future development of domain-specific complexity measures.

\subsubsection{Irreducibility Argument Sketch}\label{sec19d}

Rather than a proof, a plausibility argument is shared below:

\begin{enumerate}
    \item Let $C$ represent the full contextual history (rationale, constraints, actor assumptions).
    \item Let $O$ represent a system output derived from $C$ via lossy transformation.
    \item Assume we can recover $C$ from $O$ alone without any memory infrastructure.
    \item This implies $O$ encodes $C$, contradicting the premise that $C$ includes latent, tacit, or discarded elements.
    \item Therefore, recovery of $C$ from $O$ alone is not generally feasible.
\end{enumerate}

\noindent This supports the need for persistent memory infrastructure, especially in domains where interpretability, accountability, and revision history are critical.

\subsubsection{Partial Capture and Practical Sufficiency}\label{sec19e}

Even a partial memory capture improves system transparency and resilience. Let $C' \subset C$ be a retained subset of contextual elements, and define reconstructability $R(C)$ as the degree to which system decisions can be understood or audited. Then:

\[
R(C') > R(\emptyset)
\]
\\
\noindent even if $|C' | \ll |C|$, provided that $C'$ includes high-impact discriminators such as rejected alternatives, key assumptions, or outcome annotations.

\noindent This principle mirrors practices in explainable AI: incomplete but structured recall is more valuable than opaque decision chains. CMI, therefore, advocates for structured, bounded memory as a practical compromise between irreducibility and system tractability.

\section{Methodological Contributions}\label{sec22}

The theoretical foundations of Contextual Memory Intelligence (CMI) define primitives such as contextual entropy, insight drift, and resonance intelligence. This section introduces empirical methodologies that enable these constructs to be evaluated and applied within real-world systems. These support implementation through validation, benchmarking, and longitudinal auditing. With these in place, systems can evolve based on usage patterns, user feedback, and contextual changes.

\subsection*{Evaluation and Audit Techniques}

\subsubsection{Context Lineage Tracing}\label{sec23}
\textit{Context lineage tracing} is introduced as a structured method to reconstruct how insights were generated, transformed, and reused. Drawing inspiration from data lineage in data engineering, this method documents insight evolution from original rationale and assumptions to subsequent reinterpretation and reuse.

\noindent Lineage tracing links to the theoretical constructs defined in section~\ref{sec14}, providing empirical evidence for evaluating memory degradation or drift. Lineage tracing is necessary for auditing decision chains, identifying context loss, and restoring reasoning continuity. 

\subsubsection{Memory Utility Scoring}\label{sec24}
To extend beyond binary success/failure metrics, CMI introduces \textit{memory utility scores} that quantify the value of a memory trace over time. These scores assess how well insights support evolving decisions and conditions. Factors include:

\begin{itemize}
    \item Frequency of insight reuse in related decisions
    \item User annotations, feedback, or endorsement
    \item Alignment with updated context (semantic or temporal coherence)
    \item Detection of drift, contradiction, or misalignment
\end{itemize}

\noindent These metrics turn memory into a dynamic asset that can be optimized. They also serve as practical proxies for the constructs of drift, resonance, and irreducibility (Sections~\ref{sec14}-\ref{sec16}).

\subsubsection{Evaluation Framework for CMI Systems}\label{sec26}
A multi-dimensional evaluation framework is proposed to assess the performance of memory-aware systems. Each dimension corresponds to key design and theoretical goals, offering measurable indicators for coherence, resilience, and insight utility.

\noindent
\begin{minipage}{\textwidth}
\renewcommand{\arraystretch}{1.2}
\small
\captionof{table}{Preliminary evaluation framework for CMI system performance.}
\label{tab:cmi_eval}
\centering
\begin{tabular}{p{4.8cm}p{8.5cm}}
    \toprule
    \textbf{Dimension} & \textbf{Evaluation Criteria} \\
    \midrule
    Context Capture Completeness & Proportion of relevant rationale, alternatives, and assumptions recorded \\
    Retention Integrity & Fidelity and durability of contextual links across systems and time \\
    Regeneration Fidelity & Accuracy of reconstructed context relative to original conditions \\
    Reflection Responsiveness & Degree to which human feedback is incorporated into future reasoning \\
    Insight Utility Score & Frequency, relevance, and impact of reused memory \\
    Drift Detection Accuracy & Precision and recall of insight drift or misalignment detection \\
    \bottomrule
\end{tabular}
\end{minipage}

\vspace{1em}
\noindent This framework is designed to evaluate whether systems effectively embody the theoretical principles introduced in sections~\ref{sec13}-\ref{sec16} and support meaningful human-AI teaming as described in section~\ref{sec19f}.

\subsection*{Interoperability and Fragmentation Detection}

\subsubsection{Cross-Modal Alignment Metrics}\label{sec25}
As memory-aware systems operate across diverse modalities (e.g., dashboards, workflows, natural language, structured forms), CMI introduces \textit{cross-modal alignment metrics}. These detect semantic and structural coherence across tools and interfaces to preserve contextual continuity.

\noindent For instance, if a rationale captured in a conversational agent diverges significantly from how it is presented in a report dashboard, the system can flag this misalignment for review. This methodology supports early fragmentation detection, echoing the entropy and drift risks outlined in section~\ref{sec11}.

\FloatBarrier

\section{Contextual Deficiencies in Generative AI Systems}\label{sec19f}
Despite their remarkable fluency, large language models (LLMs) capacity to comprehend and effectively use context is still severely restricted. This shortcoming is not merely a technical constraint. Existing AI paradigms are structurally unprepared to fill this crucial gap. Agent-based systems exacerbate flaws in LLM architecture, and there will be an enduring need for human judgment grounded in computational irreducibility.

\subsection{Core Gaps in LLMs}
Despite significant progress, LLMs exhibit persistent contextual shortcomings. They often fail to retain or expose the rationale behind their outputs. They rarely preserve rejected alternatives critical in high-stakes domains like healthcare, where transparency and traceability are essential \cite{arxiv2502.15871, arxiv2505.00675}. They typically rely on shallow memory systems such as Retrieval-Augmented Generation (RAG), which depend on explicit queries and well-structured data but fall short of capturing dynamic, evolving reasoning paths \cite{memorag, arxiv2505.00675}.\\

\noindent Even with extended context windows now ranging from 128K to over 1M tokens, models gain quantitative breadth but not qualitative depth \cite{longcontextllm}. Long contexts can introduce noise and distract from salient memory cues, limiting coherence and retrieval fidelity \cite{infllm}. Moreover, reasoning encoded in model weights remains non-transparent and non-recoverable, reinforcing a brittle form of insight generation that lacks cumulative memory or explanation \cite{arxiv2505.00675, generalscales}. The result is a persistent data-to-knowledge asymmetry, in which organizations collect vast information but struggle to convert it into actionable, interpretable knowledge \cite{gaiopportunities, aisafetysurvey}.\\

\noindent Additionally, the surface-level fluency of LLM outputs creates an illusion of coherence that can mask unstable or inconsistent reasoning. Self-reflection mechanisms are externally prompted and do not represent actual memory or internal understanding \cite{iguazio}. While In-Context Learning (ICL) appears to simulate reasoning, it does so without continuity across sessions or persistent memory, states \cite{aisafetysurvey}. The majority of memory tools available today, such as those built into agent frameworks or Model Context Protocol (MCP), are designed to enhance AI task performance. More robust external memory structures like Mem0 and CrewAI are designed to enhance AI task performance \cite{chhikara2025mem0, duan2024exploration}. These systems are not designed to detect when reasoning has drifted, nor can they track or revise underlying assumptions. As such, they optimize for immediate performance at the expense of longitudinal coherence, interpretability, and adaptive reflection \cite{bubeck2023}. \\

\noindent While many existing paradigms address aspects of memory, reasoning, or task continuity, none fully support the longitudinal coherence, reflective reuse, and rationale preservation proposed by Contextual Memory Intelligence (CMI). Table \ref{tab:cmi_comparison} compares CMI to related memory and reasonign paradigms.

\clearpage
\begin{sidewaystable}
\centering
\caption{Comparison of CMI and Related Memory and Reasoning Paradigms}
\begin{tabular}{p{3.5cm} p{3.5cm} p{4.2cm} p{5.5cm}}
\toprule
\textbf{Paradigm} & \textbf{Primary Focus} & \textbf{Limitations (per CMI)} & \textbf{Key Distinctions from CMI} \\
\midrule
\textbf{Reflective AI} \newline (Chain-of-Thought, ReAct, Reflexion) & Task-level self-reflection and iterative reasoning & Lacks persistent memory or rationale reuse & CMI enables cross-session continuity, rationale preservation, and drift-aware regeneration \\
\textbf{Episodic Memory} \newline (LangGraph, MemGPT) & Session-based memory for individual agents & Memory is local and lacks versioned rationale or cross-tool insight reuse & CMI models organizational memory and supports reflection and memory scoring \\
\textbf{Cognitive Architectures} \newline (SOAR, ACT-R, OpenCog) & Simulating symbolic or hybrid human cognition & Agent-centric; not designed for collaborative or workflow-level reasoning & CMI supports sociotechnical reasoning continuity across time and tools \\
\textbf{RAG + Reasoning Pipelines} & Context retrieval for task optimization & Optimized for immediate accuracy, lacks rationale or insight traceability & CMI introduces rationale versioning and human-in-the-loop memory curation \\
\textbf{CrewAI with External Memory} & Agent collaboration with shared memory (e.g., Redis, vector stores) & Focus on task continuity; lacks rationale tracking or semantic drift repair & CMI enables insight drift detection, memory scoring, and reflective oversight \\
\textbf{Mem0} & Long-term memory for AI agents & Strong persistence but lacks structured insight reuse and human feedback loops & CMI captures \emph{why} decisions were made and supports versioned reasoning repair \\
\bottomrule
\end{tabular}
\label{tab:cmi_comparison}
\end{sidewaystable}
\clearpage

\noindent\textit{\textbf{In short, current LLMs lack persistent contextual understanding, and their fluent outputs remain disconnected from deeper reasoning memory.}}

\subsection{AI Agents and the Context Problem: A Double-Edged Sword}
AI agents promise distributed reasoning and task delegation, but without structured memory design, they risk compounding the context problem. Fragmented coordination is a common problem in multi-agent systems, due to truncated histories and unclear intent propagation. \cite{modelcontext, llmfailuremodes, securitymas, threatsurvey}. These failures are amplified when agents operate in partially observable environments or when user goals are underspecified.\\

\noindent Crucially, parametric memory does not support reconstructing missing rationale after decisions are made \cite{arxiv2505.00675}. Because of this constraint, fragmented context becomes systemic and difficult to restore. Although agent frameworks can be engineered to support persistent, shared memory and knowledge continuity \cite{llmagenticsurvey, llmreasoningsurvey}, they are simplistic frameworks primarily with the purpose of passing context. Longer context windows or chained inference offer partial mitigation but remain computationally expensive and often degrade performance due to input noise and alignment drift \cite{longcontextllm, infllm}. In short, agent collaboration without memory coherence can introduce more problems than it solves due to the scale of compounding errors and the prevalence of silent process failures. \\

\noindent\textit{\textbf{AI agents alone cannot resolve the context problem. They often intensify it unless paired with a shared memory infrastructure.}}

\subsection{The Unknowable Full Context: Human-in-the-Loop as a Necessity}

Beyond technical workarounds lies a fundamental epistemic constraint: computational irreducibility. Certain cognitive and contextual processes cannot be compressed or inferred from final outputs alone. They require replaying the complete sequence of decisions and assumptions that shaped them \cite{computationalirreducibility}. LLMs encode these reasoning pathways within opaque, distributed weights, rendering them largely unrecoverable even to the models themselves \cite{arxiv2505.00675, generalscales}.\\

\noindent This makes human oversight not merely helpful but structurally indispensable. The “centaur model” of human-AI teaming, first demonstrated in chess and now applied across domains, offers a robust framework for compensating mutual limitations \cite{kasparov, centaurevals}. Empirical evidence shows that such hybrid teams consistently outperform humans and AI alone, particularly in high-stakes, creative, and ill-defined tasks \cite{mitsynergy, hybridintelligence}. The success of these teams stems from complementary strengths: humans bring metacognition, ethical reasoning, intuition, and contextual understanding, while AI contributes to speed, precision, and pattern detection at scale \cite{metacognition, mckinsey}.\\

\noindent However, this synergy is not automatic. Even collaborative systems can fail without persistent, shared memory and mechanisms for reflection and revision. A central challenge is the “human oversight paradox,” where humans, particularly in the presence of confident AI-generated explanations, may overtrust flawed outputs \cite{fosteroversight, iguazio}. AI explanations can unintentionally suppress human critical evaluation, especially in contexts where cognitive biases like anchoring or confirmation bias dominate \cite{confidenceai, psychologytoday}.\\

\noindent To resolve this, epistemic humility must be cultivated on both sides: AI systems must acknowledge their limitations through transparency and structured memory, and humans must engage their metacognitive faculties to assess AI outputs critically \cite{epistemichumility, dell2022}. Contextual Memory Intelligence (CMI) responds to this dual need by offering an architectural foundation for collaborative intelligence. It enables systems to retain, regenerate, and trace past decisions' rationale while embedding mechanisms for human-in-the-loop correction and insight reinforcement over time. This infrastructure transforms human-AI teaming from opportunistic coordination into intentional, adaptive partnership.\\

\noindent\textit{\textbf{Systems can only achieve the trust, traceability, and reflection required for complex reasoning with human-in-the-loop design and shared memory.}}

\subsection{Summary: Contextual Deficiencies in Generative AI Systems}
The inability of current generative AI systems to preserve, reconstruct, or reason over context across time is not a peripheral flaw; it is a structural limitation. While promising, AI agents amplify this weakness without deliberate memory integration. Human oversight remains essential, yet it cannot function effectively without access to contextual history. Addressing these interconnected deficiencies requires more than incremental improvements. A comprehensive memory framework that empowers humans and agents through shared, persistent, and interpretable context is needed. Contextual Memory Intelligence offers this foundation.

\begin{tcolorbox}[colback=gray!10, colframe=black, boxrule=0.5pt]
In Summary:
\begin{itemize}
    \item Agent memory helps an AI system remember context to complete tasks.
    \item CMI helps humans and AI remember context to enhance decision-making.
\end{itemize}
\textit{\textbf{The central design question for AI systems should not be how to make them more efficient, but what problem we are trying to solve.}}\\ \\
\textbf{Key Takeaway:} Synergy between humans and AI is not automatic; both require epistemic humility, shared memory, and critical oversight. The framework needed to make this synergy scalable, resilient, and reflective is provided by CMI.
\end{tcolorbox}

\section{Empirical and Paradigmatic Justification}\label{sec19f}

This section provides empirical evidence and theoretical grounding for Contextual Memory Intelligence (CMI) as a novel and necessary paradigm. It outlines the real-world limitations of current approaches, demonstrates field-level gaps, and offers a formal argument for CMI as a paradigmatic shift in the treatment of memory within intelligent systems.

\subsection{Empirical Necessity and Practical Gaps}\label{sec27}

Despite the widespread adoption of AI tools, knowledge management platforms, and collaborative systems, organizations continue to suffer from siloed information and lost rationale. Empirical studies in organizational learning and system usability \cite{dell2022, bartek2023} highlight the high cost of contextual discontinuity, including onboarding delays, repeated mistakes, and poor decision traceability.\\

\noindent Most modern applications log what was done but do not track why. This context gap is especially acute in AI-driven workflows, where models make predictions but lack access to up-to-date information. These omissions result in significant costs to organizations.

\subsection{Gaps in Current Approaches to Contextual Retention}\label{sec28}

Several system classes attempt to manage knowledge or memory, but none resolve the deeper challenge of persistent, reflective, and cross-temporal context:

\begin{itemize}
    \item \textbf{Knowledge Management Systems} archive static documents without supporting evolving rationales or tracking insight reuse.
    \item \textbf{AI Memory Architectures} such as RAG, MemGPT, and LangGraph offer session-based persistence but lack durable, reasoning-aware memory across time and tools.
    \item \textbf{Human-Computer Interaction Models} presume that context continuity resides with the user rather than being embedded in system architecture.
\end{itemize}

CMI redefines memory not as a passive byproduct but as an active infrastructure.
\vspace{1em}
\noindent
\begin{minipage}{\textwidth}
\renewcommand{\arraystretch}{1.2}
\small
\captionof{table}{Unmet system needs addressed by Contextual Memory Intelligence (CMI).}
\label{tab:field_gaps}
\centering
\begin{tabular}{p{5.2cm}ccc}
    \toprule
    \textbf{Capability Gap} & \textbf{KM Systems} & \textbf{AI / LLM Architectures} & \textbf{CMI} \\
    \midrule
    Captures evolving rationale        & $\times$ & $\times$ & \checkmark \\
    Retains rejected alternatives      & $\times$ & $\times$ & \checkmark \\
    Tracks insight reuse over time     & $\sim$   & $\times$ & \checkmark \\
    Supports reasoning restoration     & $\times$ & $\times$ & \checkmark \\
    Enables drift-aware insight aging  & $\times$ & $\times$ & \checkmark \\
    \bottomrule
\end{tabular}

\vspace{0.5em}
\noindent\textbf{Symbol legend:} \checkmark = full support; $\sim$ = partial or indirect support; $\times$ = not supported.
\end{minipage}

\subsection{Implementation Challenges and Considerations}\label{sec30}

CMI also introduces nontrivial implementation considerations. These are best understood as design and adoption challenges that must be addressed:

\subsubsection{Technical and Regulatory Challenges}
\begin{itemize}
    \item \textit{Scalability and Storage}: Persistent memory across time, tools, and actors requires scalable indexing, decay scoring, and retrieval-based summarization to avoid performance bottlenecks.
    \item \textit{Privacy and Compliance}: CMI systems must align with GDPR, HIPAA, and ISO/IEC 42001 through role-based access controls, retention policies, and rationale redaction.
\end{itemize}

\subsubsection{Organizational and Strategic Challenges}
\begin{itemize}
    \item \textit{Cultural Adoption}: Transparent memory may surface previously implicit rationales, challenging power dynamics, and accountability structures. Implementation requires role-based memory visibility and participatory governance.
    \item \textit{Cost-Benefit Dynamics}: Initial deployment may carry technical and procedural overhead. Empirical pilots are necessary to demonstrate ROI over traditional KM or audit systems.
\end{itemize}

\noindent These limitations reinforce the need for responsible design and interdisciplinary evaluation, not a retreat from innovation.

\subsection{Paradigm Shift: From Retrieval to Reasoning Infrastructure}\label{sec31}

CMI proposes a paradigmatic reorientation: Memory is no longer an archival artifact but an infrastructural element as essential as inference or data governance. Existing systems view memory through the lens of \textit{retrieval}; CMI reframes it as a dynamic mechanism for sustaining contextual reasoning over time.

\subsubsection{From Static Content to Dynamic Infrastructure}
While traditional systems treat information as objects, CMI draws from situated and distributed cognition to conceptualize memory as a durable substrate that enables reuse, reinterpretation, and resilience across workflows and timescales.

\subsubsection{Theoretical Distinctiveness}
While Contextual Memory Intelligence (CMI) draws inspiration from existing paradigms such as audit trails, data provenance, version control, and context-aware systems, it departs significantly in scope and intent. CMI reframes these constructs through a unified architectural lens, treating memory not as a byproduct of documentation or runtime state but as a persistent, interpretable infrastructure for reflective reasoning. Table~\ref{tab:theoretical_distinctiveness} summarizes how CMI reenvisions each adjacent area.

\noindent
\begin{minipage}{\textwidth}
\renewcommand{\arraystretch}{1.2}
\small  
\captionof{table}{Theoretical distinctiveness of Contextual Memory Intelligence (CMI) compared to adjacent paradigms.}
\label{tab:theoretical_distinctiveness}
\centering
\begin{tabular}{p{5.2cm}p{8.2cm}}  
    \toprule
    \textbf{Comparison Dimension} & \textbf{How CMI Differs} \\
    \midrule
    Audit Trails vs. Reflective Reasoning & CMI enables evolving rationale capture, not just static compliance records. \\
    Data Provenance vs. Interpretive Context & CMI extends provenance to include human assumptions, rejected options, and decision intent. \\
    Version Control vs. Semantic Drift Detection & CMI models how meanings evolve over time, not just changes in content. \\
    Context-Aware Systems vs. Contextual Continuity & CMI treats context as a persistent, reconstructable structure, not just a runtime parameter. \\
    \bottomrule
\end{tabular}
\end{minipage}

\subsection{Evaluating CMI as a Paradigm Shift}\label{sec33a}
To further support the claim that Contextual Memory Intelligence (CMI) represents a paradigmatic shift, this section maps CMI against Thomas Kuhn’s criteria for scientific revolutions. These include novelty, incommensurability, and superior problem-solving capacity. CMI satisfies all core dimensions, as indicated in Table~\ref{tab:kuhn_cmi}.

\FloatBarrier  

\begin{table}[t]
\centering
\renewcommand{\arraystretch}{1.3}
\caption{Evaluating CMI Against Kuhn’s Paradigm Shift Criteria}
\label{tab:kuhn_cmi}
\begin{tabular}{p{4cm}p{9.5cm}}
    \toprule
    \textbf{Kuhn Criterion} & \textbf{CMI Alignment} \\
    \midrule
    \textbf{Novelty} & CMI introduces original constructs such as contextual entropy, insight drift, resonance intelligence, and computational irreducibility, which are absent in existing AI, KM, and HCI paradigms. \\
    
    \textbf{Incommensurability} & CMI redefines memory as infrastructure, not metadata. This ontological shift renders current paradigms conceptually incompatible with memory-aware reasoning. \\
    
    \textbf{Problem-Solving Power} & CMI directly addresses persistent problems such as context fragmentation, rationale loss, drift detection, and lack of cross-temporal coherence in intelligent systems. \\
    
    \textbf{Conceptual Coherence} & CMI integrates multiple constructs under a unified architectural and theoretical framework, offering internal consistency across use cases and domains. \\
    
    \textbf{Generativity} & CMI enables new research areas, including drift epidemiology, memory utility scoring, resonance modeling, and context regeneration pipelines. \\
    
    \textbf{Community Reorientation} & CMI suggests new evaluation criteria, design practices, and system roles (e.g., memory stewards) that reorient how intelligent systems are built and assessed. \\
    
    \textbf{Anomaly Resolution} & CMI resolves longstanding gaps in audit trails, provenance tracking, and explainability by preserving evolving context and enabling reflective insight reuse. \\
    \bottomrule
\end{tabular}
\end{table}

\FloatBarrier

\subsection{Generative Potential for New Research Programs}\label{sec33}

CMI enables new methodological and empirical pathways that are not easily absorbed by current paradigms. These include:

\begin{itemize}
    \item \textbf{Insight Drift Detection} - Measuring semantic misalignment across time.
    \item \textbf{Memory Utility Scoring} - Ranking retained insights by coherence, reuse, and impact.
    \item \textbf{Context Regeneration Models} - Reconstructing decision logic under changing conditions.
    \item \textbf{Reflective System Design} - Architectures that incorporate human-in-the-loop memory repair.
\end{itemize}

\noindent These programs do not merely extend existing lines of research. They instantiate a new logic of memory-centered intelligence.

\subsection{Architectural and Organizational Implications}\label{sec34}

A paradigm shift must also produce systemic consequences. CMI reframes organizational memory as an active capability, prompting changes in many areas. Four example areas are governance, multi-agent systems, strategic planning, and institutional learning. CMI supports the area of governance by increasing traceability and auditability. It supports Multi-Agent Systems by enabling persistent memory across interacting agents. It supports strategic planning by retaining historical rationale for long-range alignment. It also supports institutional learning by capturing outputs and the context behind them. These shifts demonstrate how CMI transforms memory from static recordkeeping into a generative infrastructure that reshapes how institutions, teams, and systems think, adapt, and collaborate. \\

\noindent\textit{\textbf{CMI transforms memory from an afterthought into a foundation essential for trust, reasoning, and adaptive intelligence.}}

\subsubsection{Interoperability: CrewAI, MCP, and CMI}
While emerging agent frameworks like CrewAI and Model Context Protocol (MCP) improve AI task coordination and continuity, they focus primarily on enabling efficient task execution. In contrast, Contextual Memory Intelligence (CMI) addresses a more profound need: shared memory infrastructures that support long-term insight reuse, rationale preservation, and human-AI collaboration. Below is an example to show how they can complement each other to form powerful workflows when combined. \\ \\
\textbf{Use Case: AI Agents Supporting Customer Escalation Reviews} \\ \\

\noindent In this scenario, a set of AI agents supports a customer success team by analyzing escalated complaints, summarizing case history, identifying root causes, suggesting responses, and recommending policy improvements. 

\vspace{1em}
\textbf{Layered Roles and Responsibilities}
\begin{itemize}
    \item \textbf{CrewAI with External Memory:} Agents carry out specialized tasks with access to shared external memory (vector databases, Redis, document stores). This memory enables agents to persist local task data such as past tickets, retrieved policy documents, and inter-agent outputs.
    
    \item \textbf{Model Context Protocol (MCP):} MCP enables context standardization and continuity by passing structured memory (chat history, tool outputs, retrieved documents) into LLM calls in a uniform format. It helps link conversations, task history, and tool usage across agent chains and sessions. In this use case, MCP orchestrates handoffs between agents while retaining user-facing task context.

    \item \textbf{Contextual Memory Intelligence (CMI):} CMI overlays both CrewAI and MCP to provide reflective context tracking. It captures the “why” behind decisions, logs rationale and rejected alternatives, tracks long-term insight reuse, and supports human-in-the-loop review. CMI transforms operational context into adaptive memory for decision support and collaboration.
\end{itemize}

\vspace{1em}
\textbf{Use Case Workflow}
\begin{enumerate}
    \item CrewAI agents handle a customer complaint:
    \begin{itemize}
        \item Agent 1 summarizes prior tickets.
        \item Agent 2 retrieves relevant policy documents.
        \item Agent 3 drafts a resolution proposal.
        \item Agent 4 compares similar past cases using external memory.
    \end{itemize}
    \item External memory stores task-specific information, such as previous case data, outputs, and retrieval logs.
    \item CMI tracks higher-order patterns and rationale:
    \begin{itemize}
        \item Recognizes trending complaint categories over time.
        \item Stores rationale behind rejected draft responses.
        \item Links decisions to long-term outcomes (e.g., reduced churn, satisfaction trends).
        \item Recommends reusing or adapting previously successful resolution strategies.
    \end{itemize}
\end{enumerate}

\vspace{1em}

\section{Conclusion and Future Work}\label{sec36}
Contextual Memory Intelligence (CMI), a new multidisciplinary field that allows systems to remember and reason with changing contexts over time, was presented in this paper. In contrast to conventional approaches that treat memory as static, CMI reconceptualizes memory as an adaptive infrastructure essential to organizational learning and decision continuity.

\subsection{Summary of Contributions}\label{sec37}
This literature made four primary contributions:
\begin{itemize}
    \item \textbf{Theoretical} - Defined CMI as a shift from "information as object" to "memory as infrastructure" and introduced primitives such as insight drift, contextual entropy, and resonance intelligence.
    \item \textbf{Architectural} - Proposed the Insight Layer, a modular system framework for embedding contextual memory into human-AI workflows.
    \item \textbf{Methodological} - Outlined novel techniques, including context lineage tracing, memory utility scoring, and cross-modal alignment metrics.
    \item \textbf{Field Positioning} - Established CMI as an emerging field of study, offering integrative models and system architectures that bridge theoretical gaps across AI, knowledge management, and human-computer interaction.
\end{itemize}

\subsection{Future Research Directions}\label{sec38}
The emergence of CMI creates several options for future inquiry:
\begin{itemize}
    \item \textbf{Empirical Validation} - Development of benchmark datasets and longitudinal studies to assess memory utility, drift detection, and fidelity.
    \item \textbf{Human-AI Interfaces} - Designing interactive environments where users can reflect on, revise, and contribute to persistent memory structures.
    \item \textbf{Integrations} - Embedding CMI capabilities in enterprise tools, LLM-based agents, CRM systems, and collaborative platforms.
    \item \textbf{Memory Ethics and Governance} - Investigating the ethical implications of preserving organizational context, including power dynamics, transparency, and memory redaction policies.
\end{itemize}

\subsection{Toward Reflective and Resilient Systems}\label{sec39}
The inability to preserve and recall context represents a critical gap that must be addressed as organizations integrate AI-augmented workflows and agents that augment or automate complex decision-making; AI systems risk reinforcing shallow decisions and undermining institutional knowledge without structured memory. CMI provides a pathway toward reflective, auditable, and resilient systems that can explain what was done, why, how it evolved, and what should be done. Through this, CMI overcomes organizational and technical constraints and establishes the groundwork for a more coherent, equitable, and adaptive future for intelligent systems.

\section*{Declarations}\label{sec52}
\subsection*{Funding}
Not applicable.

\subsection*{Conflict of interest}
The author declares no conflict of interest.

\clearpage



\end{document}